\documentclass[sigconf]{aamas}

\usepackage{balance} 
\usepackage{comment}
\usepackage{hyperref}
\usepackage{xurl}  

\doi{INST9866}

\makeatletter
\gdef\@copyrightpermission{
  \begin{minipage}{0.2\columnwidth}
   \href{https://creativecommons.org/licenses/by/4.0/}{\includegraphics[width=0.90\textwidth]{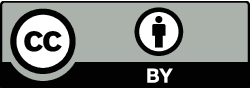}}
  \end{minipage}\hfill
  \begin{minipage}{0.8\columnwidth}
   \href{https://creativecommons.org/licenses/by/4.0/}{This work is licensed under a Creative Commons Attribution International 4.0 License.}
  \end{minipage}
  \vspace{5pt}
}
\makeatother

\setcopyright{ifaamas}
\acmConference[AAMAS '26]{Proc.\@ of the 25th International Conference
on Autonomous Agents and Multiagent Systems (AAMAS 2026)}{May 25 -- 29, 2026}
{Paphos, Cyprus}{C.~Amato, L.~Dennis, V.~Mascardi, J.~Thangarajah (eds.)}
\copyrightyear{2026}
\acmYear{2026}
\acmDOI{}
\acmPrice{}
\acmISBN{}

\title[Title]{TriBand-BEV: Real-Time LiDAR-Only 3D Pedestrian Detection via Height-Aware BEV and High-Resolution Feature Fusion}

\author{Mohammad Khoshkdahan}
\affiliation{
  \institution{Karlsruhe Institute of Technology}
  \city{Karlsruhe}
  \country{Germany}}
\email{mohammad.khoshkdahan@kit.edu}

\author{Alexey Vinel}
\affiliation{
  \institution{Karlsruhe Institute of Technology}
  \city{Karlsruhe}
  \country{Germany}}
\email{alexey.vinel@kit.edu}

\begin{abstract}
Safe autonomous agents and mobile robots need fast real time 3D perception, especially for vulnerable road users (VRUs) such as pedestrians. We introduce a new bird's eye view (BEV) encoding, which maps the full 3D LiDAR point cloud into a light-weight 2D BEV tensor with three height bands. We explicitly reformulate 3D detection as a 2D detection problem and then reconstruct 3D boxes from the BEV outputs. A single network detects cars, pedestrians, and cyclists in one pass. The backbone uses area attention at deep stages, a hierarchical bidirectional neck over P1 to P4 fuses context and detail, and the head predicts oriented boxes with distribution focal learning for side offsets and a rotated IoU loss. Training applies a small vertical re bin and a mild reflectance jitter in channel space to resist memorization. We use an interquartile range (IQR) filter to remove noisy and outlier LiDAR points during 3D reconstruction.

On KITTI dataset, TriBand-BEV attains 58.7/52.6/47.2 pedestrian BEV AP(\%) for easy, moderate, and hard at 49 FPS on a single consumer GPU, surpassing Complex-YOLO, with gains of +12.6\%, +7.5\%, and +3.1\%. Qualitative scenes show stable detection under occlusion. The pipeline is compact and ready for real time robotic deployment. Our source code is publicly available on GitHub.\footnote{\url{https://github.com/mohammadkhsh/TriBand-BEV}}

\end{abstract}


\newcommand{\BibTeX}{\rm B\kern-.05em{\sc i\kern-.025em b}\kern-.08em\TeX}

\begin{document}


\pagestyle{fancy}
\fancyhead{}


\maketitle 


\section{Introduction}
Mobile robots and autonomous vehicles (AVs) rely on 360-degree, real-time environmental perception to safely navigate and interact within complex outdoor spaces. Meeting this fundamental requirement demands sensors capable of high-frame-rate feedback and reliable operation in diverse environmental conditions, all while adhering to stringent compute and power budgets. While cameras offer rich semantic and appearance data, achieving full 360-degree coverage necessitates the use of multiple units \cite{nguyen2022multi}, which increases the computational footprint, and their performance is further compromised by factors like glare, fog, and low light. Moreover, recent studies report subgroup biases in camera-based pedestrian detection related to orientation, occlusion, and appearance attributes \cite{khoshkdahan2025fair, khoshkdahan2025beyond}. Radar provides weather-robust data, but its inherent angular coarseness makes it unsuitable for achieving the sub-meter precision object bounding box and orientation necessary for high-fidelity 3D object detection at large distances \cite{li2022exploiting}. Consequently, LiDAR has emerged as the principal sensor, offering superior range accuracy and stable geometric measurements. However, the resulting large, irregular, and sparse point cloud data introduces considerable challenges related to memory consumption and computational complexity, which prevents efficient deployment on embedded robotic platforms \cite{wu2020deep}.

\begin{figure}[t]
    \vspace{0.2cm}
    \centering
    \includegraphics[width=1\linewidth]{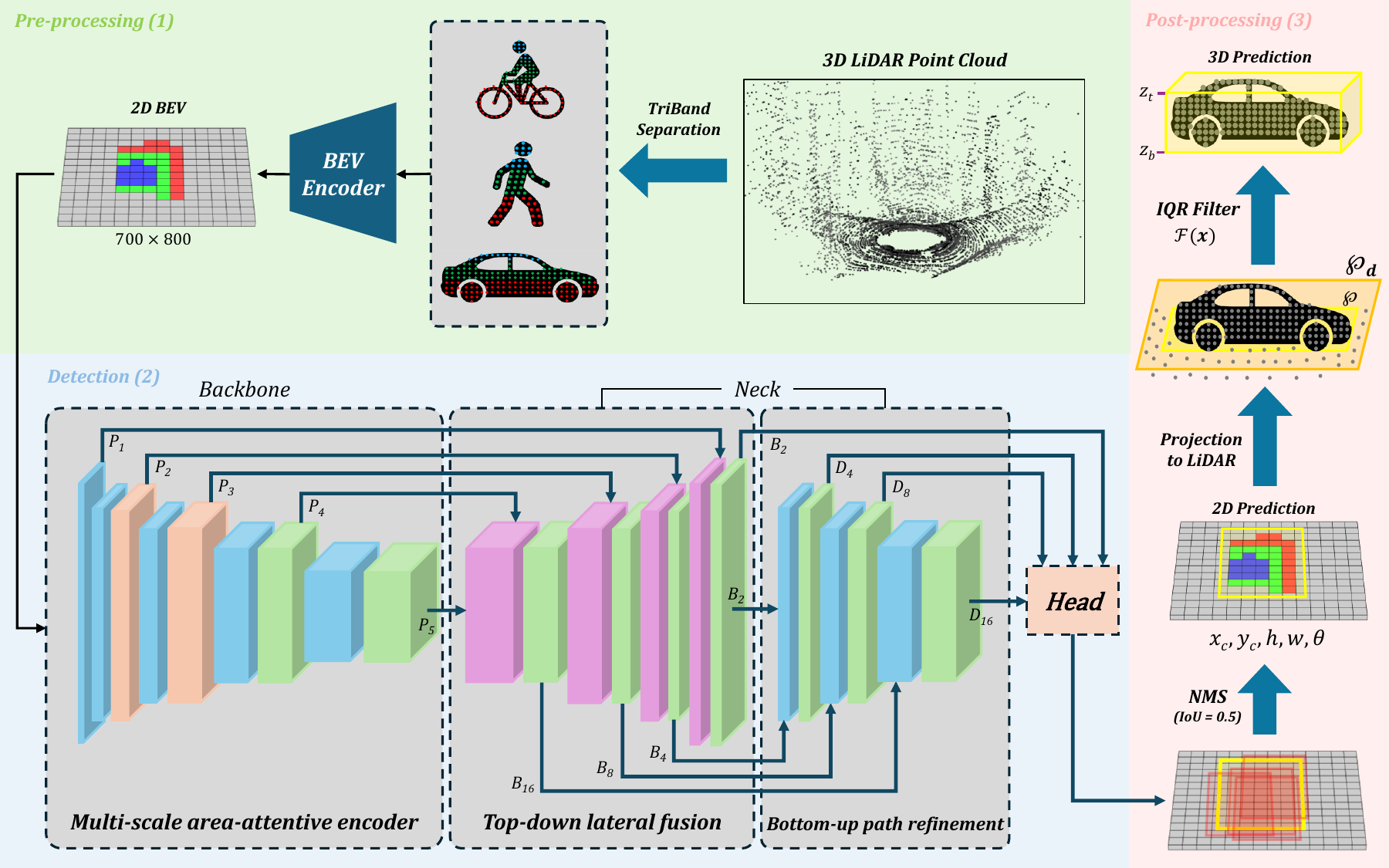}
    \caption{Detection pipeline overview. The pre-processing part (first step) converts 3D points to our novel BEV encoding and the detection part (second step) produces the 2D predictions on the BEV map. Lastly, post-processing (third step) generates the final 3D predictions. }
    \label{fig:intro}
\end{figure}

Existing LiDAR processing pipelines fall into a few broad families, namely voxelization \cite{zhou2018voxelnet, deng2021voxelrcnn} in 3D grids, point-based processing of neighborhoods \cite{qi2017pointnet, shi2019pointrcnn, qi2017pointnetplusplus}, fusion with cameras \cite{pang2020clocs, sindagi2019mvxnet}, sparse convolution in 3D \cite{yan2018second}, and projection to 2D views \cite{zhou2019fvnet, meyer2019lasernet, simony2018complex} such as bird's eye view (BEV) or front view (FV). We specifically focus on the BEV approach due to its ability to map the scene into a compact 2D tensor (800×700 pixels) with only three channels, which constitutes a highly compressed data representation, and enables fast inference. This property is especially important for embedded robots and traffic moderation tasks. Importantly, the main goal of this study is to advance real-time, lightweight 3D detection rather than to surpass all state-of-the-art methods, and to significantly improve upon Complex-YOLO \cite{simony2018complex} within the BEV domain. To this end, we introduce the following core contributions.

Our first contribution is TriBand-BEV Encoding, a novel projection scheme that transforms the 3D point cloud into a three-channel BEV map. Each channel aggregates the maximum reflectance within a distinct height band, which preserves coarse vertical structure and object identity. Our second contribution is to processes the TriBand representation using a multi-resolution backbone enhanced with an efficient area attention mechanism to capture regional context more effectively than conventional convolutions. Standard feature pyramid networks (FPNs) include very low-resolution maps to cover large object scales in images, but in BEV all objects appear similar in footprint irrespective of range, and coarse maps lose critical spatial detail for small BEV boxes. Our bidirectional neck avoids these low-resolution features and instead uses an expanded set of high-resolution maps that the detection head processes, which enhances accuracy for small objects. The next contribution is about the training augmentation. We apply channel-space perturbations, specifically reflectance jittering and height re-binning (shift), to enhance model generalization and robustness against the different object elevations inherent in real-world LiDAR data. Finally, for stable 3D box recovery, we employ a Fast IQR-Based Post-Processing Filter that stabilizes height estimation by removing outlier noise and yields robust 3D reconstruction. The full pipeline, illustrated in Fig.~\ref{fig:intro}, keeps the compute profile close to that of 2D detectors while retaining the geometric benefits of LiDAR.

\section{Related Works}

\subsection{Multi-Modal Fusion Approaches}
Multi-sensor 3D detection leverages camera imagery for rich semantics and LiDAR for precise geometry. 
Early-fusion models such as F-PointNet \cite{qi2018frustum} and F-ConvNet \cite{wang2019frustum} inject 2D detections into frustum-constrained point clouds, which reduces the 3D search space. While boosting accuracy, these pipelines are tightly coupled to image proposals, so missed objects in the camera view propagate as missed detections in 3D. PointPainting \cite{vora2020pointpainting} enriches each LiDAR point with pixel-wise semantic masks, but misalignment between views and increased input dimensionality limit efficiency.  

Intermediate-fusion approaches merge features mid-network. MVX-Net \cite{sindagi2019mvxnet} indexes image features into voxelized LiDAR grids, while ContFuse \cite{liang2018contfuse} projects image features into BEV using continuous convolution operators. More advanced designs such as 3D-CVF \cite{yoo20203dcvf}, EPNet \cite{huang2020epnet}, and Transformer-based methods like TransFusion \cite{bai2022transfusion} use cross-attention to adaptively select image cues for LiDAR features. These fine-grained interactions improve accuracy but add computational overhead and require cross-view alignment.  

Late-fusion frameworks such as CLOCs \cite{pang2020clocs} combine outputs from independent 2D and 3D detectors. This avoids cross-modal calibration at the feature level and is robust to sensor failures, but it sacrifices deep semantic integration and runs two detectors in parallel, which again increases the inference time.  

Attention-based fusion is a prominent recent direction. Cross-modal Transformers, e.g., TransFusion \cite{bai2022transfusion}, as well as multi-modal attention schemes like mmFUSION \cite{ahmad2023mmfusion}, dynamically align features across modalities, producing strong results on benchmarks. However, these methods demand high GPU memory and careful temporal synchronization.

\subsection{LiDAR-Only Detection}
LiDAR-only pipelines avoid calibration overhead and focus purely on geometric data. Their differences stem from how the 3D point cloud is represented.

Projection-based methods map LiDAR into 2D views for efficient detection. Range-view projections (e.g., Velo-FCN \cite{li2016vehicle}, FVNet \cite{zhou2019fvnet}, LaserNet \cite{meyer2019lasernet}) map point clouds into dense cylindrical images and apply 2D CNNs. LaserNet predicts probabilistic bounding boxes per point, clustering them to generate 3D detections. However, range images suffer from occlusion and long-range sparsity. BEV projection, introduced in MV3D \cite{chen2017mv3d} and refined in PIXOR \cite{yang2018pixor}, preserves ground-plane geometry and simplifies yaw estimation. However, the PIXOR approach discretizes the height range of (-2.5,1)m into 35 vertical slices plus one reflectance channel, which results in a deep, computationally heavy 36-channel input tensor (800×700×36). This input structure mirrors a high-resolution 3D voxelization collapsed onto the BEV plane, which again creates a processing overhead that limits real-time speed on resource-constrained platforms. Extensions like FaF \cite{luo2018fast} and HDMapNet \cite{li2021hdmapnet} incorporate temporal cues or map priors.

Complex-YOLO\cite{simony2018complex} converts LiDAR into a compact BEV RGB map whose three channels encode maximum height, maximum intensity, and normalized point density per cell, then applies a single-stage detector with an Euler RPN that regresses orientation via a complex-angle parameterization for efficient, real-time 3D box prediction. So our direction is to keep the real-time, BEV-only spirit but redesign the encoding and the feature processing. Instead of a point-density channel, TriBand-BEV encodes maximum reflectance across three height bands. A density channel only counts points and loses information below the highest return, whereas multi-band reflectance preserves vertical structure that is more informative and easier for a detector to learn. The detector operates on a bidirectional, high-resolution multi-scale fusion to strengthen small-object cues while maintaining the throughput expected of BEV pipelines. 

Voxel-based methods discretize space into 3D grids and learn features via convolution. VoxelNet \cite{zhou2018voxelnet} introduced end-to-end voxel feature encoding (VFE) via PointNets inside each voxel, replacing hand-crafted features. SECOND \cite{yan2018second} improved efficiency with sparse convolutions, skipping empty voxels to achieve real-time throughput. Voxel R-CNN \cite{deng2021voxelrcnn} added voxel RoI pooling for proposal refinement, boosting localization of small objects. Recent designs like SST \cite{fan2022single} and Voxel Transformer \cite{mao2021voxeltransformer} embed self-attention into voxel backbones to enlarge the receptive field. These models preserve 3D structure but inference cost grows cubically with voxel resolution, challenging deployment at fine granularity.  

Point-based methods directly operate on raw LiDAR points with learned neighborhood aggregation. PointNet \cite{qi2017pointnet} and PointNet++ \cite{qi2017pointnetplusplus} initiated direct learning on raw points through set abstraction. PointRCNN \cite{shi2019pointrcnn} segments foreground points and generates bottom-up proposals, achieving high accuracy but with large computational demand. STD \cite{yang2019std} and 3DSSD \cite{yang20203dssd} proposed anchor-free single-stage pipelines with distance-aware point sampling to handle far and sparse objects. Graph and attention operators, as in Point-GNN \cite{shi2020pointgnn} and Pointformer \cite{pan20213d}, capture non-local dependencies, while PV-RCNN \cite{shi2020pvrcnn} combines voxel backbones with point-level keypoint refinement. These methods excel in accuracy but point-level MLP and neighborhood queries scale poorly to dense scenes, making real-time inference difficult.  

Hybrid methods combine voxel efficiency with point precision. PV-RCNN++ \cite{shi2021pvrcnn++} and Pyramid R-CNN \cite{mao2021pyramidrcnn} perform voxel encoding followed by point-level refinement across multiple resolutions. Range-based hybrids like LaserFlow \cite{meyer2021laserflow} leverage temporal range images for lightweight joint detection and motion forecasting. Hybrids often achieve top benchmarks but incur latency from voxel–point transformations and added architectural complexity.  

\subsection{Discussion}

Fusion methods often sit at the top of leaderboards because they blend appearance and geometry, but they come with higher latency, larger memory footprints, and a constant need for careful calibration between sensors. LiDAR-only detectors remain appealing for mobile robots and embedded platforms since they are simpler to deploy, less fragile to calibration drift, and easy to maintain in the field. The bottleneck is computation. Many strong 3D systems process large point sets or voxel grids and the model size is often between 50M to over 120M parameters. Even recent state-of-the-art methods report non-trivial runtimes: BFT3D ($\sim180$ ms)~\cite{bft3d}, CasA ($64-114$ ms)~\cite{casa}, ELPF-FM ($9.8$ FPS)~\cite{elpf-fm}, Fade3D ($12$ FPS)\cite{fade3d}, TED-S ($90$ ms)~\cite{teds}, ViKIENet-R ($15$ FPS)~\cite{vikienet}, and VirConv ($\sim52$ ms)~\cite{virconv}.

A compact BEV tensor addresses this constraint. With only three channels, a single frame occupies on the order of tens of kilobytes, far below the megabyte scale of raw points and well below typical voxel representations. This reduction enables fast inference and modest memory use, which are both essential for onboard perception. Absolutely, a performance gap against fusion or full 3D volumetric pipelines is expected because those families benefit from richer cues or denser spatial reasoning, but they pay for it with speed and complexity.

\section{Dataset and Preprocessing}
\label{sec:data}

\subsection{KITTI Dataset}
\label{sec:data:kitti}

Experiments are conducted on the KITTI dataset~\cite{geiger2012kitti}, a widely used benchmark for 3D object detection in autonomous driving. Data were collected in Karlsruhe, Germany, using a Velodyne HDL\mbox{-}64E LiDAR scanner with 64 beams and a $360^\circ$ horizontal field of view. The 3D detection track includes 7{,}481 annotated frames and 7{,}518 test frames without released labels. Performance is evaluated using average precision (AP), which summarizes the precision–recall curve into a single detection score.

Following common practice, we adopt a half–half split of the available labeled data (official training set), using 3{,}712 samples for training and 3{,}769 for validation. The split is constructed in a sequence–disjoint manner, so that training and validation frames originate from different driving sequences, in line with established KITTI partition strategies~\cite{chen20153d, chen2017mv3d}. In contrast, random splitting (as done in some earlier studies such as~\cite{shao2023efficient}) leads to shared scenes between the two sets. This allows the model to memorize parts of the environment, resulting in unrealistically high scores (e.g., car BEV AP above 98\% in the easy subset with our model) that do not reflect true generalization to unseen data.

KITTI defines three evaluation difficulty levels namely, easy, moderate, and hard, based on 2D bounding box height, occlusion state, and truncation ratio. 
Specifically, objects are categorized according to minimum bounding box height (40, 25, 25 pixels), maximum occlusion level (0, 1, 2), and maximum truncation (15\%, 30\%, 50\%), respectively. 
This design provides a stratified assessment of detectors under favorable, common, and highly challenging conditions.  

In our work, only LiDAR data are used as model input for training and evaluation and RGB images are employed solely for qualitative visualization of detection outputs.

\subsection{Bird’s Eye View (BEV) Encoding}
\label{sec:data:bev}

We rasterize the ROI $x\!\in\![0,70]$\,m and $y\!\in\![-40,40]$\,m into $\Delta\!=\!0.1$\,m cells ($W\!=\!700$, $H\!=\!800$). For each cell $(u,v)$, let $\mathcal{C}_{u,v}$ denote the set of LiDAR returns in that cell. Heights are expressed in meters relative to a reference plane located $1.73$\,m below the sensor, corresponding to the ground beneath the ego vehicle. The returns are partitioned into three vertical bands: $\mathcal{B}_1{:}\;z<0.65$\,m, $\mathcal{B}_2{:}\;0.65\le z<1.30$\,m, and $\mathcal{B}_3{:}\;z\ge1.30$\,m. 

Some returns exhibit very low reflectance values due to surface material properties, incidence angle effects, or partial energy absorption, which can make them underrepresented after discretization. To avoid losing these measurements in the BEV image, we add $0.1$ to each reflectance and amplify it by $30\%$. Let $\tilde{\rho}_i = 1.3\,(\rho_i + 0.1)$ denote the corrected reflectance of return $i$. The BEV image encodes, per cell and per band, the maximum corrected reflectance, scaled to 8-bit:

\begin{equation}
I_k(u,v) \;=\; 255 \times 
\max_{\,i \in \mathcal{C}_{u,v} \cap \mathcal{B}_k}\;\tilde{\rho}_i
\quad\text{for } k\in\{1,2,3\},
\end{equation}
with $I_k(u,v){=}0$ if the set is empty. The final 3-channel BEV is
\begin{equation}
\mathbf{I}(u,v) \;=\; \big[I_1(u,v),\,I_2(u,v),\,I_3(u,v)\big],
\end{equation}
which we map to $(R,G,B)$ respectively. This three-band design preserves coarse vertical structure after the 3D$\to$2D projection. Rather than collapsing each cell to a single height cue, it encodes distinct vertical regions. For pedestrians, legs mainly activate the lower band, torso and arms the middle band, and head the upper band. For vehicles, wheels and bumper dominate the lower band, body panels the middle band, and the roof the upper band. This layered response yields a more informative BEV pattern and improves separability.

\section{Network Architecture}
\subsection{Overview}
\label{sec:method:overview}

\begin{figure}[t]
  \centering
  \includegraphics[width=\linewidth]{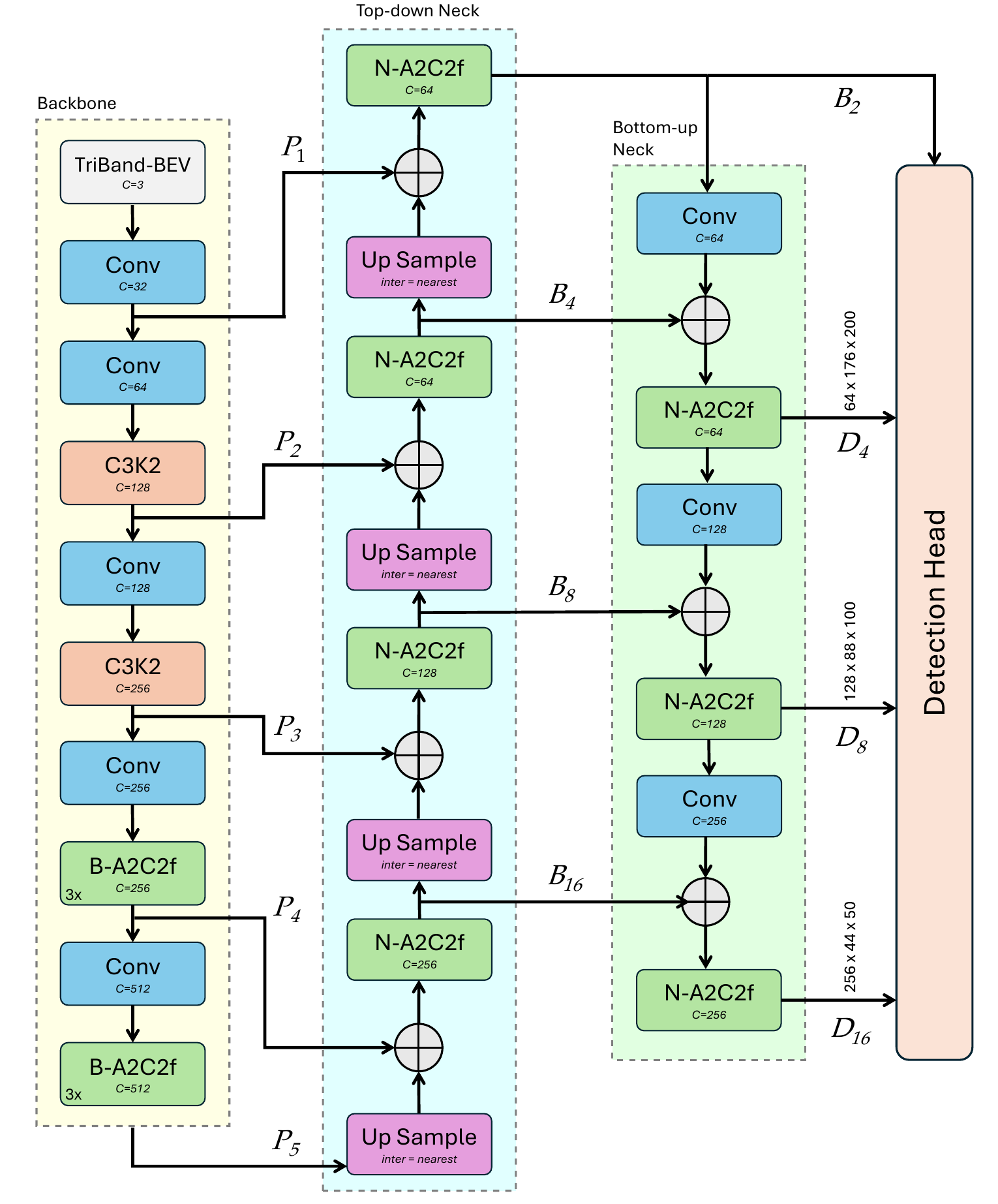}
  \caption{Network overview. The detection pipeline from the BEV input (3 channels) through the backbone and two-stage bidirectional neck to the head. Each block is annotated with $C{=}\,x$ indicating the number of output channels for that feature map. The head predicts, at each spatial location, oriented BEV box parameters (center offsets, width, length, yaw), class logits, and an objectness confidence score.}
  \label{fig:arch}
\end{figure}

The proposed architecture (see Fig. \ref{fig:arch}) divides into backbone, neck, and head components. The backbone transforms the BEV input \(I \in \mathbb{R}^{3 \times H \times W}\) into a pyramid of feature maps \(\{P_1, P_2, P_3, P_4, P_5\}\), where \(P_1\) retains the finest spatial resolution and \(P_5\) the coarsest. The modules placed in backbone instances include \texttt{C3k2} \cite{khanam2024yolov11} for maintaining spatial detail and \texttt{A2C2f} (introduced by \cite{tian2025yolov12}) for integrating area attention and contextual features.   

The neck implements dual feature-fusion pathways. A top-down path upsamples deeper, semantically rich feature maps toward higher resolution levels and merges them via lateral connections with corresponding backbone features. A bottom-up refinement path downsamples fused high-resolution maps and merges them with coarse backbone maps in order to sharpen object localization, especially of small objects.

We modify the architecture proposed by \cite{tian2025yolov12} and extend it by enlarging both directions and fusing high resolution feature maps. The head now uses fused feature levels, denoted \(\mathcal{F} = \{B_2, D_4, D_8, D_{16}\}\), for detection. The highest resolution detection level corresponds to half the input image resolution. This contributes to higher recall for objects whose BEV projections cover few pixels (such as pedestrians). The head outputs class probabilities, center offsets, size, and orientation for each fused level.

\begin{figure}[t]
  \centering
  \includegraphics[width=0.95\linewidth]{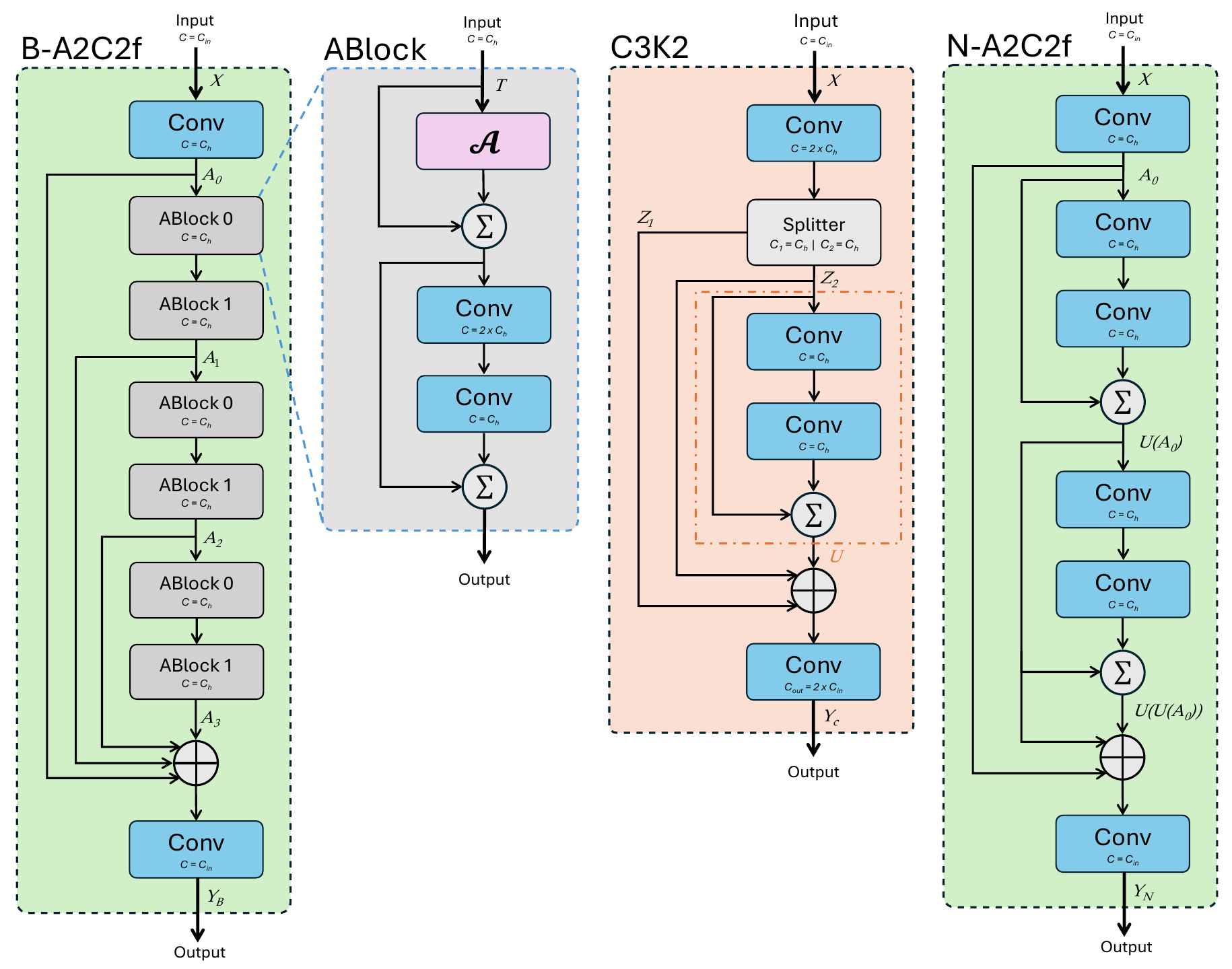}
  \caption{Internal components of each module used in the proposed architecture.}
  \label{fig:modules}
\end{figure}

\subsection{Backbone and Building Blocks}
\label{sec:method:backbone}

We use two families of modules in the backbone. In this section we explain the details of the \texttt{C3k2} and \texttt{B-A2C2f} block (see Fig.~\ref{fig:modules}). Throughout, let
\begin{equation}
\phi_{k\times k,\,s}^{(c_{\mathrm{in}}\!\rightarrow c_{\mathrm{out}})}:\ 
\mathbb{R}^{B\times c_{\mathrm{in}}\times H\times W}\ \longrightarrow\ 
\mathbb{R}^{B\times c_{\mathrm{out}}\times \tfrac{H}{s}\times \tfrac{W}{s}},
\end{equation}
denote a convolution of kernel size $k$ and stride $s$, and let $\phi_{1\times1}$ be a $1\times1$ convolution used for channel fusion with $s=1$. Channel concatenation is written as $\oplus$.

In addition to the specialized blocks described below, the backbone and the second stage of the neck employ standard stride-2 convolutions of the form $\phi_{3\times3,,2}^{(c_{\mathrm{in}}\rightarrow c_{\mathrm{out}})}$, progressively reducing the spatial resolution to build a multi-scale hierarchical representation.

\subsubsection{C3k2 (with a single residual Bottleneck)}

Given an input feature map defined as \(X\in\mathbb{R}^{B\times C_{\mathrm{in}}\times H\times W}\) and target width \(C_{\mathrm{out}}\), define \(C_h=\lfloor e\,C_{\mathrm{out}}\rfloor\) with \(e=0.5\). A \(1{\times}1\) convolution expands the channels to \(2C_h\) and then splits them evenly:
\begin{equation}
[Z_1,Z_2]=\mathrm{Split}\!\left(\phi_{1\times 1}^{(C_{\mathrm{in}}\rightarrow 2C_h)}(X);\ 2\right),
\qquad Z_1,Z_2\in\mathbb{R}^{B\times C_h\times H\times W}.
\end{equation}
The refinement stream applies a single bottleneck with residual addition:
\begin{equation}
U(Z_2) = Z_2+\phi_{3\times 3}^{(C_h\rightarrow C_h)}\!\Big[\phi_{3\times 3}^{(C_h\rightarrow C_h)}(Z_2)\Big].
\end{equation}
Finally, cross–stage aggregation concatenates all paths and projects to \(C_{\mathrm{out}}\):
\begin{equation}
Y_c=\phi_{1\times 1}^{(3C_h\rightarrow C_{\mathrm{out}})}\!\Big(Z_1 \oplus \ Z_2 \oplus\ U(Z_2)\Big).
\end{equation}

This block combines an identity-preserving split with a residual refinement stream inside a compact bottleneck, followed by channel fusion. The design balances gradient flow, spatial fidelity, and nonlinear transformation.

\subsubsection{B-A2C2f: Area-Attention Module}

The feature maps are processed by a specialized variant of the C2f scaffold, here denoted \texttt{B-A2C2f}. This module enhances feature processing by stacking attention--augmented residual units while preserving a shortcut branch. Let the input be \(X \in \mathbb{R}^{B\times C_{\mathrm{in}}\times H\times W}\) and the target output width be \(C_{\mathrm{out}}\). We define the hidden width as \(C_h=\lfloor e\,C_{\mathrm{out}}\rfloor\) with \(e=0.5\). 

First, a \(1{\times}1\) convolution compresses the input channels to the hidden dimension:
\begin{equation}
A_{0} = \phi_{1\times 1}^{(C_{\mathrm{in}}\rightarrow C_h)}(X).
\end{equation}

The module then applies three refinement stages. Based on the architecture design, each stage \(S(\cdot)\) comprises a sequence of two stacked ABlocks. Let \(A_{i}\) denote the output of the \(i\)-th stage:
\begin{equation}
A_{i} = S(A_{i-1}) = \mathrm{ABlock}\Big(\mathrm{ABlock}(A_{i-1})\Big), \qquad i=1, 2, 3.
\end{equation}

The outputs of the initial projection and all stages are aggregated by concatenation and fused by a final \(1{\times}1\) convolution to restore the target channel width:
\begin{equation}
Y_{\mathrm{B}} = \phi_{1\times 1}^{\left(4C_h \rightarrow C_{\mathrm{out}}\right)} \left(
A_{0} \oplus A_{1} \oplus A_{2} \oplus A_{3}
\right).
\end{equation}

Each \texttt{ABlock} integrates an area--based attention operator \(\mathcal{A}(\cdot)\) and a channel feedforward network. The operator \(\mathcal{A}\) partitions the feature map into disjoint spatial regions, forms query–key–value triplets, and applies multi–head attention within each region. This design reduces quadratic complexity while retaining long–range dependencies. The feedforward part expands the channels by factor \(\rho=2\) and compresses them back. Given \(T \in \mathbb{R}^{B\times C_h\times H\times W}\),
\begin{equation}
\mathrm{ABlock}(T) = T \;+\; \mathcal{A}(T) \;+\;
\phi_{1\times 1}^{(\rho C_h\rightarrow C_h)} \!\Big[
\phi_{1\times 1}^{(C_h\rightarrow \rho C_h)}\!\big(T+\mathcal{A}(T)\big)
\Big].
\end{equation}

The B-A2C2f module thus maintains a dense gradient flow by concatenating the raw projection \(A_0\) with the refined features from the three internal stages.

\subsection{Bi-Directional Multi-Resolution Neck}
\label{sec:method:fpn}

The neck aggregates multi-scale features through a bi-directional pyramid. A top–down stream propagates semantics from coarse backbone levels toward finer resolutions, while a bottom–up stream refines coarse scales by injecting high-resolution detail. This design ensures that both shallow and deep cues contribute to the detection stages.

Let $\mathrm{Up}_2(\cdot)$ denote $2\times$ nearest-neighbor upsampling and $\phi_{3\times 3,\,2}^{(C\!\rightarrow C)}(\cdot)$ a stride–2 convolution. At each fusion point, an \texttt{N-A2C2f} block consolidates inputs. For the top–down pathway,
\begin{equation}
B_{2^i} = \mathrm{N\!-\!A2C2f}\ \Big(
\mathrm{Up}_2(B_{2^{i+1}}) \;\oplus\; P_{i+1}
\Big), \quad i=1,2,3,
\end{equation}
where the initialization $B_{16}$ follows the same formula with $P_5$ as the upsampled seed.  

The bottom–up refinement path then propagates fine detail upward:
\begin{equation}
D_{2^{i+1}} = \mathrm{N\!-\!A2C2f}\ \Big(
\phi_{3\times 3,\,2}^{(C\!\rightarrow C)}(D_{2^i}) \;\oplus\; B_{2^{i+1}}
\Big), \quad i=2,4,
\end{equation}
with $D_4$ initialized analogously from $B_2$ and $B_4$. Detection operates on the fused set
\(\mathcal{F} = \{B_{2},D_{4},D_{8},D_{16}\}\),
balancing high-resolution precision and semantically enriched context.

This \texttt{N-A2C2f} mirrors the scaffold of the backbone’s \texttt{B-A2C2f} but replaces attention blocks with lightweight bottlenecks. As in the backbone, a $1{\times}1$ projection first reduces channel width and a final $1{\times}1$ convolution fuses concatenated states. The refinement is realized by two bottlenecks applied sequentially. Combining all steps into one expression,
\begin{equation}
Y_{N}=\phi_{1\times 1}^{(3C_h\rightarrow C_{\mathrm{out}})}\!\Big(
A_{0} \;\oplus\; U(A_{0}) \;\oplus\; U\!\big(U(A_{0})\big)
\Big),
\end{equation}
where $A_{0}=\phi_{1\times 1}^{(C_{\mathrm{in}}\rightarrow C_h)}(X)$ is the initial projection and $U(\cdot)$ is the bottleneck operator defined earlier.  

By consolidating $A_0$ with its successive bottleneck refinements, the \texttt{N-A2C2f} achieves efficient multi-path fusion while keeping computational cost low, complementing the heavier attention-based backbone blocks.

\subsection{Head \& Loss Function}
\label{sec:method:head-loss}

The head outputs per fused level: side‐distance distributions, class logits, and angle logits. Side distributions are converted to continuous offsets via DFL. The final output per cell is decoded into oriented boxes in BEV plus class score.

The training objective is

\begin{equation}
\mathcal{L}
= 7.5 \;\mathcal{L}_{\mathrm{box}}
\;+\; 1.5 \;\mathcal{L}_{\mathrm{DFL}}
\;+\; 0.5 \;\mathcal{L}_{\mathrm{cls}}.
\end{equation}

\(\mathcal{L}_{\mathrm{box}}\) is the rotated IoU loss between predicted oriented bounding boxes and ground truth.  
\(\mathcal{L}_{\mathrm{DFL}}\) is the discrete‐to‐continuous distance loss over side distributions (via Distribution Focal Loss).  
\(\mathcal{L}_{\mathrm{cls}}\) is the classification loss (sigmoid cross‐entropy) over object classes.

\subsection{3D Box Recovery from BEV}
\label{sec:method:recovery}

\begin{figure}
    \centering
    \includegraphics[width=1\linewidth]{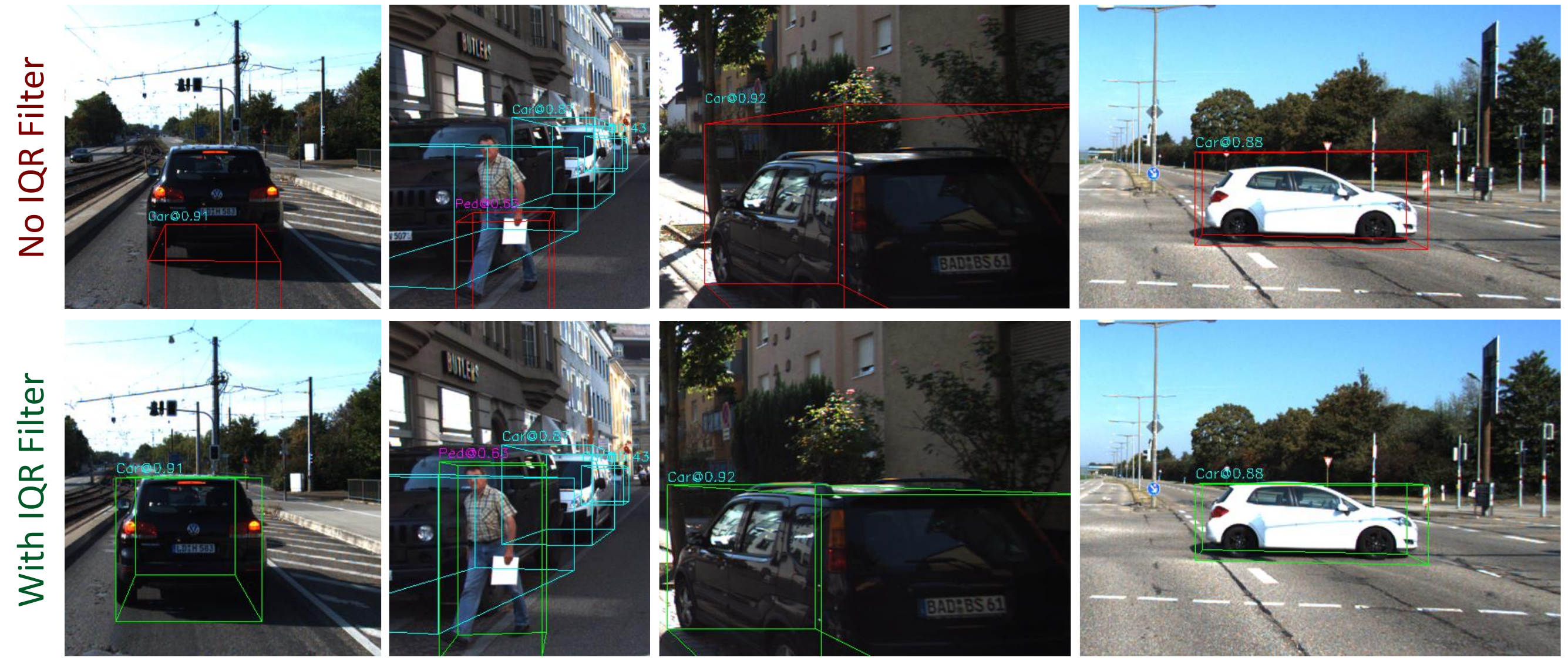}
    \caption{Effect of IQR filtering on 3D box reconstruction. Top row shows boxes reconstructed without IQR filtering. Bottom row shows results after filtering. In the first two examples (from left), low outlier returns within the same footprint shift the estimated bottom plane downward, leading to excessive height and activation of the default height constraint, which reduces 3D IoU. IQR removes these low outliers and restores a correct bottom estimate. In the third and fourth examples, the bottom plane is correct but a few high outliers inflate the top estimate. After IQR filtering, the top plane aligns with the vehicle roof, resulting in a tighter and more accurate 3D box.}
    \label{fig:iqr-filter}
\end{figure}

We convert each BEV prediction to a KITTI-format 3D box by lifting the 2D footprint to the LiDAR frame and then mapping it to the rectified camera frame. The BEV corners are de-normalized to meters to form an oriented polygon $\mathcal{P}$ in LiDAR $(x,y)$; its yaw and planar center define the footprint. To improve bottom-plane estimation at long range (sparser sweeps), we apply a distance-adaptive isotropic dilation
\begin{equation}
\mathcal{P}_{d} \;=\; s(d)\,\mathcal{P}, 
\qquad
s(d)=1+\alpha\,\frac{d}{d_{\max}}\;(\alpha\!=\!2.5,\;d_{\max}\!=\!80\text{ m}),
\end{equation}
and collect LiDAR points inside $\mathcal{P}_{d}$ for the bottom query and inside $\mathcal{P}$ for the top query.

Let $Z_{\downarrow}$ be the 10 smallest $z$ values and $Z_{\uparrow}$ the 10 largest $z$ values among the selected points. We suppress outliers with an IQR filter. Briefly, for a vector $x$, $Q_1(x)$ and $Q_3(x)$ denote the 25th and 75th percentiles and $\mathrm{IQR}(x)=Q_3(x)-Q_1(x)$. We keep inliers via the Tukey fence
\begin{equation}
\mathcal{F}(x)=\big\{\,x_i \;\big|\; Q_{1}(x)-1.5\,\mathrm{IQR}(x)\le x_i \le Q_{3}(x)+1.5\,\mathrm{IQR}(x)\,\big\}.
\end{equation}

We then take the bottom and top as $z_{\mathrm{b}}=\min \mathcal{F}(Z_{\downarrow})$ and $z_{\mathrm{t}}=\max \mathcal{F}(Z_{\uparrow})$. Figure~\ref{fig:iqr-filter} illustrates the effect of IQR filtering on 3D box reconstruction. In practice, spurious LiDAR returns may appear within the BEV footprint due to multipath reflections, small clutter, or reflective surfaces. These points can lie significantly below the true ground contact or above the actual object surface and distort the estimated bottom or top planes. In addition, at longer ranges the reduced point density may lead to insufficient samples near the true object extremities, which can further destabilize height estimation. To enforce geometric plausibility, we constrain the final height. If the top to bottom difference is below 1.25,m or above 2.1,m, we interpret this as either missing structural points or contamination by vertical outliers and assign a default height of 1.6,m above the estimated bottom. The combination of IQR filtering and this height prior yields more stable and physically consistent 3D boxes.


\begin{figure}[t]
  \centering
  \includegraphics[width=\linewidth]{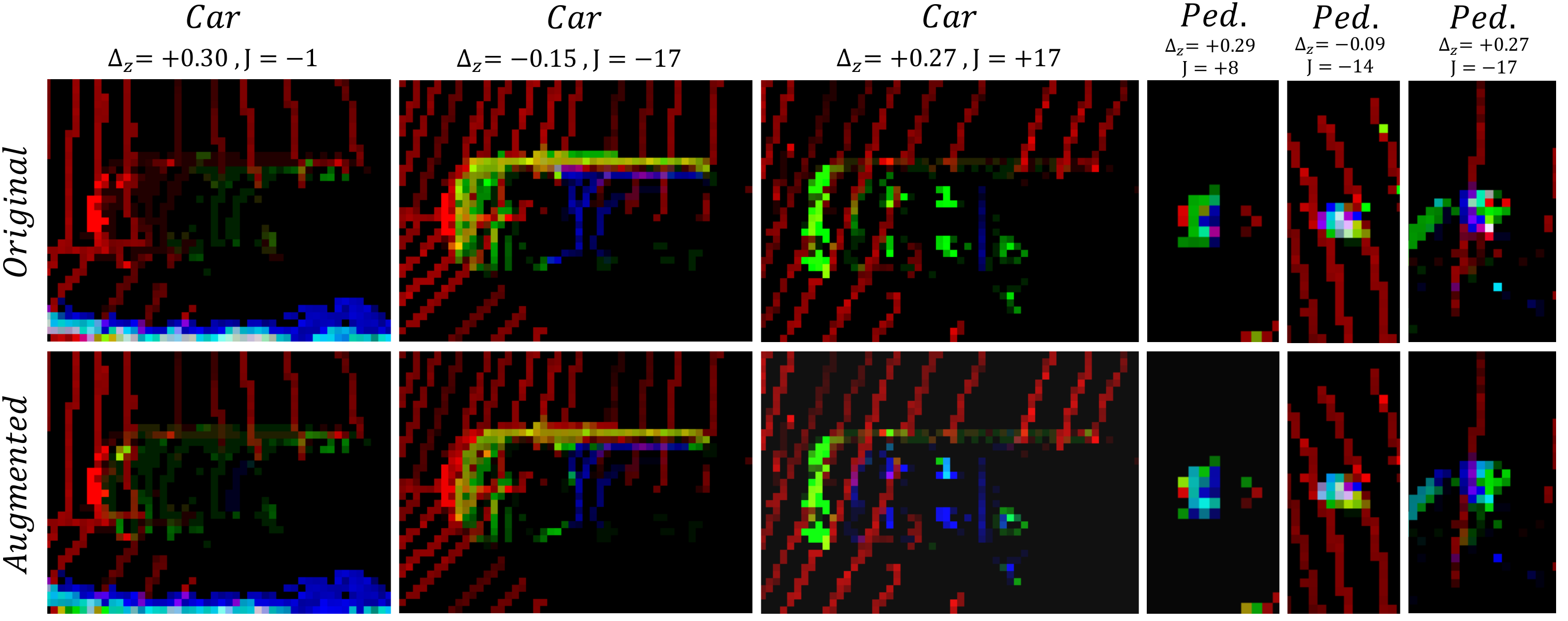}
  \caption{Random crops from the training set before (top) and after augmentation (bottom). A small vertical re-binning ($\Delta z$) shifts the activation across height bands. For example, when a car lies on a sloped surface and its returns fall mainly into the red and green bands, a positive shift can elevate the roof responses into the blue band, restoring the expected vertical pattern. Conversely, a negative shift can compress activations into fewer bands. By exposing the network to such variations during training, the model learns to remain invariant to local ground offsets and preserves consistent detection performance under height perturbations.}
  \label{fig:aug}
\end{figure}

\section{Data Augmentation}
\label{sec:aug}

To reduce memorization and increase diversity, we upsample the training set by generating one perturbed BEV per scene. We sample a vertical offset $\Delta z \sim \mathcal{U}(-0.3,0.3)$\,m and apply it before BEV construction to re–bin heights. After re–binning, we add a single zero–mean Gaussian jitter $J \sim \mathcal{N}(0,\sigma^2)$ (with $\sigma=20$ in RGB units) uniformly to all nonzero pixels and then saturate to the $[0,255]$ range. Figure~\ref{fig:aug} provides intuition.

\begin{equation}
\tilde{\mathbf{I}}(u,v)
=
\mathrm{sat}_{[0,255]}
\Big(
\mathbf{I}^{\Delta z}(u,v)
+
\boldsymbol{\eta}(u,v)\,J
\Big),
\end{equation}

where $\mathbf{I}^{\Delta z}(u,v)$ is the TriBand-BEV obtained from $z' = z + \Delta z$ and $\boldsymbol{\eta}(u,v)=\mathbf{1}\{\mathbf{I}^{\Delta z}(u,v)>0\}$ masks nonzero pixels. We also tested i.i.d.\ pixel-wise noise, but it reduced AP by breaking stable reflectance associations within objects; the image-wide jitter preserves these cues while still increasing variability.

\section{Experiments}
\label{sec:exp}

\subsection{Implementation Details}
\label{sec:exp:impl}

All trainings and validation were conducted on one NVIDIA RTX 4090 Laptop GPU (16\,GB memory) using PyTorch 2.5.1 with CUDA 12.1. Training employed distributed data parallelism with automatic mixed precision (AMP). Mini-batches of 32 were processed per iteration.

The model was trained for 60 epochs. The optimizer was stochastic gradient descent (SGD) with learning rate $0.01$, momentum $0.9$, and weight decay $5\!\times\!10^{-4}$. A linear warmup of three epochs was followed by cosine annealing of the learning rate. During training we applied non maximum suppression (NMS) with an IoU threshold of $0.7$, and at inference we used an IoU threshold of $0.5$.
The average runtime for our full model (8.8M parameters) per BEV frame is 20.4\,ms, corresponding to a processing rate of 49\,FPS.

\subsection{Ablation Study}
\label{sec:exp:ablation}

We study how capacity, augmentation, and feature resolution affect performance. Capacity is controlled by \(C_{\text{base}}\), the number of output channels in the first backbone convolution; since widths double after each downsampling stage, \(C_{\text{base}}\) sets the overall model width. The baseline uses \(C_{\text{base}}{=}16\) with a shallow neck and the head consumes only three fused levels \((P_5,P_4,P_3)\) and has no access to \((P_2,P_1)\). We then add augmentation that includes the same vertical re–bin and adds a single Gaussian offset to all nonzero pixels in the image, which helps to preserve within–object reflectance patterns. Finally, we scale capacity to \(C_{\text{base}}{=}32\) and enable a high–resolution neck so that \((P_2,P_1)\) enter the fusion path (full model for our TriBand-BEV).

\begin{table}[t]
\centering
\caption{Ablation study on the validation set. Metrics are mAP, the mean AP over easy/moderate/hard in \%. \(C_{\text{base}}\) is the channel width at the highest–resolution stage, Aug.\ is the augmentation, and \emph{Head levels} list the fused feature maps provided to the head. The final two rows keep augmentation and show the effect of capacity and then high–resolution fusion.}
\label{tab:ablation_final_ieee}
\scalebox{0.72}{
\begin{tabular}{ccccccccc} 
\toprule
\multicolumn{3}{c}{\textbf{Method}}  & \multicolumn{2}{c}{\textbf{Car mAP}} & \multicolumn{2}{c}{\textbf{Pedestrian mAP}} & \multicolumn{2}{c}{\textbf{Cyclist mAP}} \\
\cmidrule(lr){1-3}\cmidrule(lr){4-5}\cmidrule(lr){6-7}\cmidrule(lr){8-9}
$\mathbf{C_{\text{base}}}$ & $\mathbf{Aug.}$ & Head Levels & BEV & 3D & BEV & 3D & BEV & 3D \\
\midrule
16 & $\times$          & $(D_{32},D_{16},B_8)$   & 70.32 & 50.97 & 29.87 & 25.81 & 31.94 & 26.38 \\
16 & $\checkmark$        & $(D_{32},D_{16},B_8)$   & 76.19 & 57.20 & 39.47 & 29.64 & 26.51 & 21.52 \\
32 & $\checkmark$        & $(D_{32},D_{16},B_8)$   & 76.26 & 57.46 & 37.94 & 28.83 & 37.96 & 32.83 \\
32 & $\checkmark$        & $(D_{16},D_8,D_4,B_2)$   & \textbf{76.58} & \textbf{58.30} & \textbf{52.89} & \textbf{40.79} & \textbf{46.24}& \textbf{41.68} \\
\bottomrule
\end{tabular}}
\end{table}

Table~\ref{tab:ablation_final_ieee} shows that increasing capacity from \(C_{\text{base}}{=}16\) to \(32\) at the same head levels \((D_{32},D_{16},B_8)\) improves car by \(+\ \!0.07\%\) BEV and \(+\ \!0.26\%\) 3D, and cyclist by \(+\ \!11.45\%\) BEV and \(+\ \!11.31\%\) 3D. However, it leads to slight performance drop for pedestrians. Enabling the high–resolution neck and moving the head to \((B_2,D_4,D_8,D_{16})\) on the same \(C_{\text{base}}{=}32\) model yields the largest additional gains (compared to same capacity): pedestrian \(+\ \!14.95\%\) BEV and \(+\ \!11.96\%\) 3D, cyclist \(+\ \!8.28\%\) BEV and \(+\ \!8.85\%\) 3D, and car \(+\ \!0.32\%\) BEV and \(+\ \!0.84\%\) 3D. 

Gains are largest for pedestrians and cyclists, while cars improve less since their BEV footprints already carry ample information after downsampling. The key challenge is small object size in the BEV grid rather than distance dependent scale as in camera images. A pedestrian may occupy only a few cells at a 0.1 m resolution, which demands high spatial fidelity. Injecting higher resolution features into the fusion path preserves these fine cues and improves both recall and localization.

To further assess robustness to the fixed height-band assumption, we apply a multi-offset inference strategy on the final configuration ($C_{\text{base}}{=}32$ with augmentation and full high-resolution fusion). At test time, three BEV encodings are generated using vertical offsets of $-0.3\,\mathrm{m}$, $0\,\mathrm{m}$, and $+0.3\,\mathrm{m}$, and predictions are merged via NMS at IoU $0.5$. Compared to the single-offset full model, this strategy yields only marginal BEV gains for pedestrians ($+0.67\%$) and cyclists ($+1.09\%$), while slightly reducing Car BEV performance ($-0.10\%$). Since inference cost increases approximately threefold, these results indicate that the proposed single-offset formulation already provides sufficient robustness with substantially higher efficiency.

\begin{figure}[t]
    \centering
    \includegraphics[width=\linewidth]{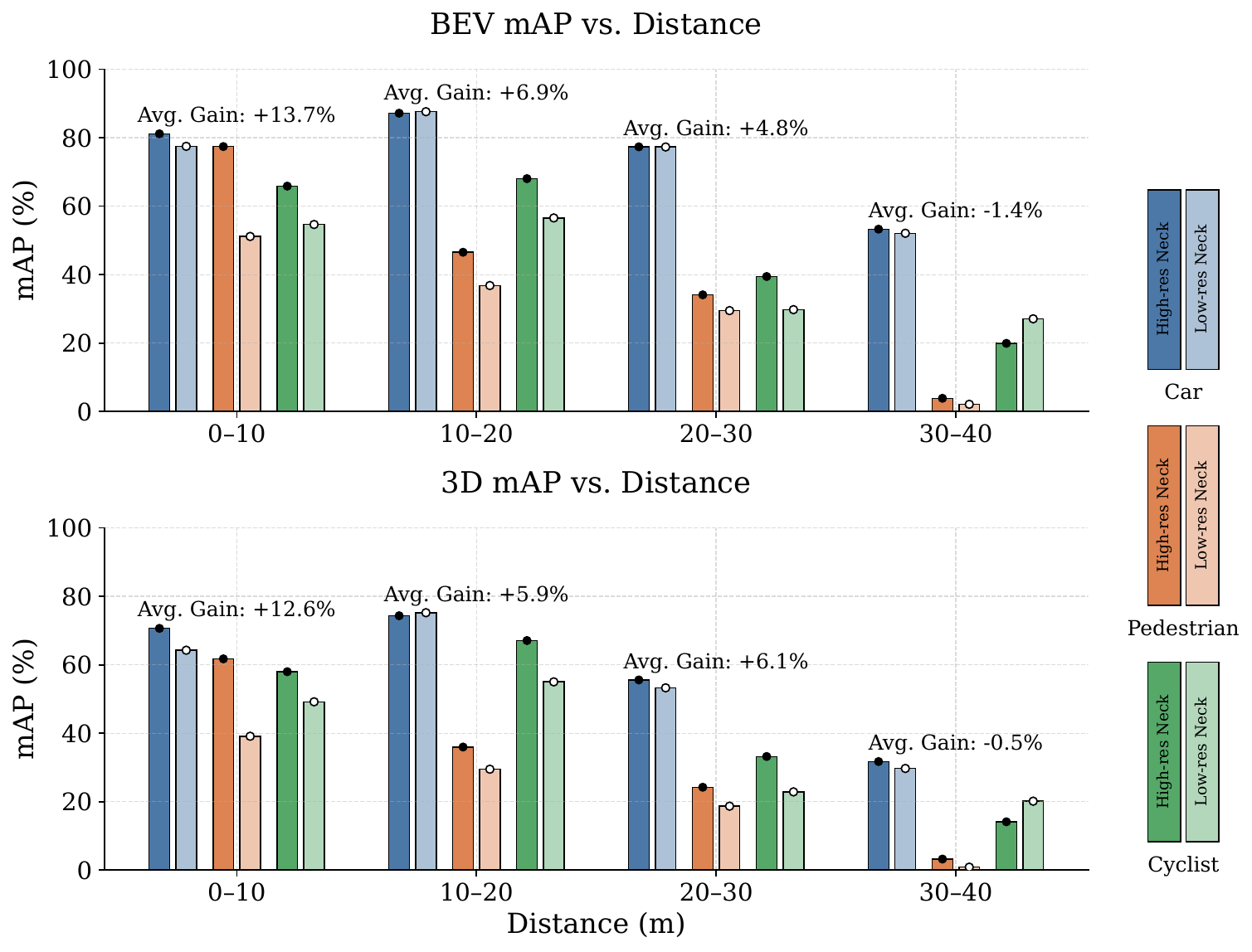}
    \caption{Mean BEV and 3D AP across distance ranges. Each group shows class-wise mAP for cars, pedestrians, and cyclists, with the mean value annotated above.}
    \label{fig:map_by_distance}
\end{figure}

We also analyze performance by distance bands and report the mean over easy, moderate, and hard for each class in Fig.~\ref{fig:map_by_distance}. Car remains more robust, whereas pedestrian and cyclist decline more beyond $30\,\mathrm{m}$. The main factor is lower LiDAR point density at distance, which reduces multi band activation in the BEV map and affects recall for smaller objects, while larger cars retain denser coverage and therefore higher accuracy.

\section{Results and Comparison}
\label{sec:exp:results}

We report 3D and BEV KITTI results with a focus on pedestrian and include inference speed for runtime comparison. Evaluation follows the official KITTI protocol with 40 recall positions and class-specific IoU thresholds (car 0.7, pedestrian 0.5, cyclist 0.5). AP is computed from the interpolated precision–recall curve over $\mathcal{R}=\{0,\tfrac{1}{39},\ldots,1\}$; at each sampled recall $r$ we use the standard monotonic interpolation:
\begin{equation}
\mathrm{AP}\;=\;\frac{1}{|\mathcal{R}|}\sum_{r\in\mathcal{R}}\;\max_{\tilde r\ge r}\;\frac{\mathrm{TP}(\tilde r)}{\mathrm{TP}(\tilde r)+\mathrm{FP}(\tilde r)}.
\end{equation}

\subsection{Quantitative Results}

\begin{figure}[t]
\centering
\includegraphics[width=1.0\linewidth]{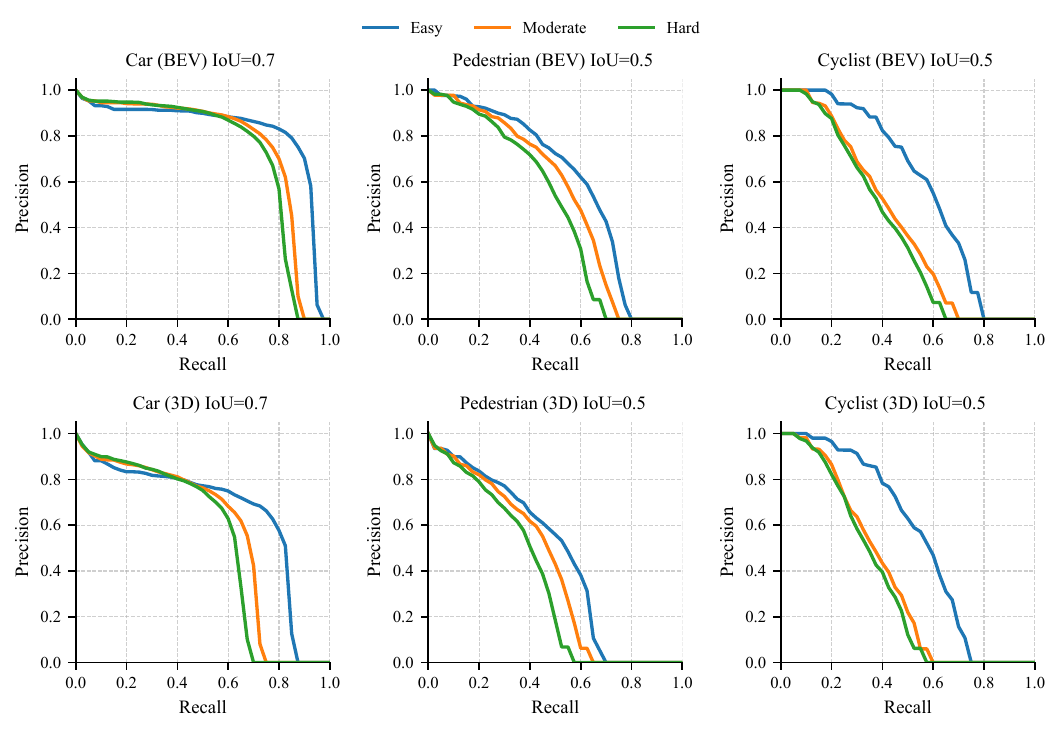}
\caption{3D and BEV precision–recall curves for easy/moderate/hard on all three classes.}
\label{fig:ped3d_pr}
\end{figure}

Table~\ref{tab:ped3d} summarizes BEV and 3D pedestrian AP by our full scale model compared to Complex-YOLO \cite{simony2018complex}. Our method attains 58.72\% / 52.68\% / 47.27\% BEV AP (easy/moderate/hard) at 49 FPS, surpassing the results by Complex-YOLO. The gains are 12.64\% (easy), +7.59\% (moderate), and +3.07\%(hard). Furthermore, our model surpassed 3D pedestrian APs for the easy and moderate difficulty. Figure~\ref{fig:ped3d_pr} shows the corresponding precision–recall curves under the same 40-point interpolation. In addition, Table~\ref{tab:bev_all} reports BEV and 3D AP for car and cyclist using identical splits.

\begin{table}[t]
\centering
\caption{3D pedestrian AP in percentage (easy/moderate/hard) at 40 recall points and FPS.}
\label{tab:ped3d}
\setlength{\tabcolsep}{2pt}
\scalebox{0.89}{
\begin{tabular}{lccc ccc c}
\toprule
Method & \multicolumn{3}{c}{\textbf{BEV AP@0.5}} & \multicolumn{3}{c}{\textbf{3D AP@0.5}} & \textbf{FPS} \\
\cmidrule(lr){2-4}\cmidrule(lr){5-7}
& Easy & Moder. & Hard & Easy & Moder. & Hard & \\
\midrule

Complex-YOLO & 46.08 & 45.09 & 44.20 & 41.79 & 39.70  & \textbf{35.92} & \textbf{50} \\
\midrule
\textbf{Ours (TriBand-BEV)} & \textbf{58.72} & \textbf{52.68} & \textbf{47.27} & \textbf{45.91} & \textbf{40.83} & 35.64 & 49\\
\bottomrule
\end{tabular}}
\end{table}

\begin{table}[t]
\centering
\caption{BEV and 3D average precision (AP) on KITTI (40 recall points) for other two common classes.}
\label{tab:bev_all}
\setlength{\tabcolsep}{6pt}
\begin{tabular}{lccc ccc}
\toprule
& \multicolumn{3}{c}{\textbf{Car AP@0.7}} & \multicolumn{3}{c}{\textbf{Cyclist AP@0.5}} \\
\cmidrule(lr){2-4}\cmidrule(lr){5-7}
Space & Easy & Mod & Hard &  Easy & Mod & Hard \\
\midrule
\textbf{BEV} &81.74 & 75.42 & 72.57 & 57.90 & 41.77 &39.04\\
\textbf{3D}  &  65.32& 56.58& 52.99 & 54.00& 36.54& 34.50\\
\bottomrule
\end{tabular}
\end{table}

\subsection{Qualitative Analysis}

\begin{figure}[t]
  \centering
  \includegraphics[width=\linewidth]{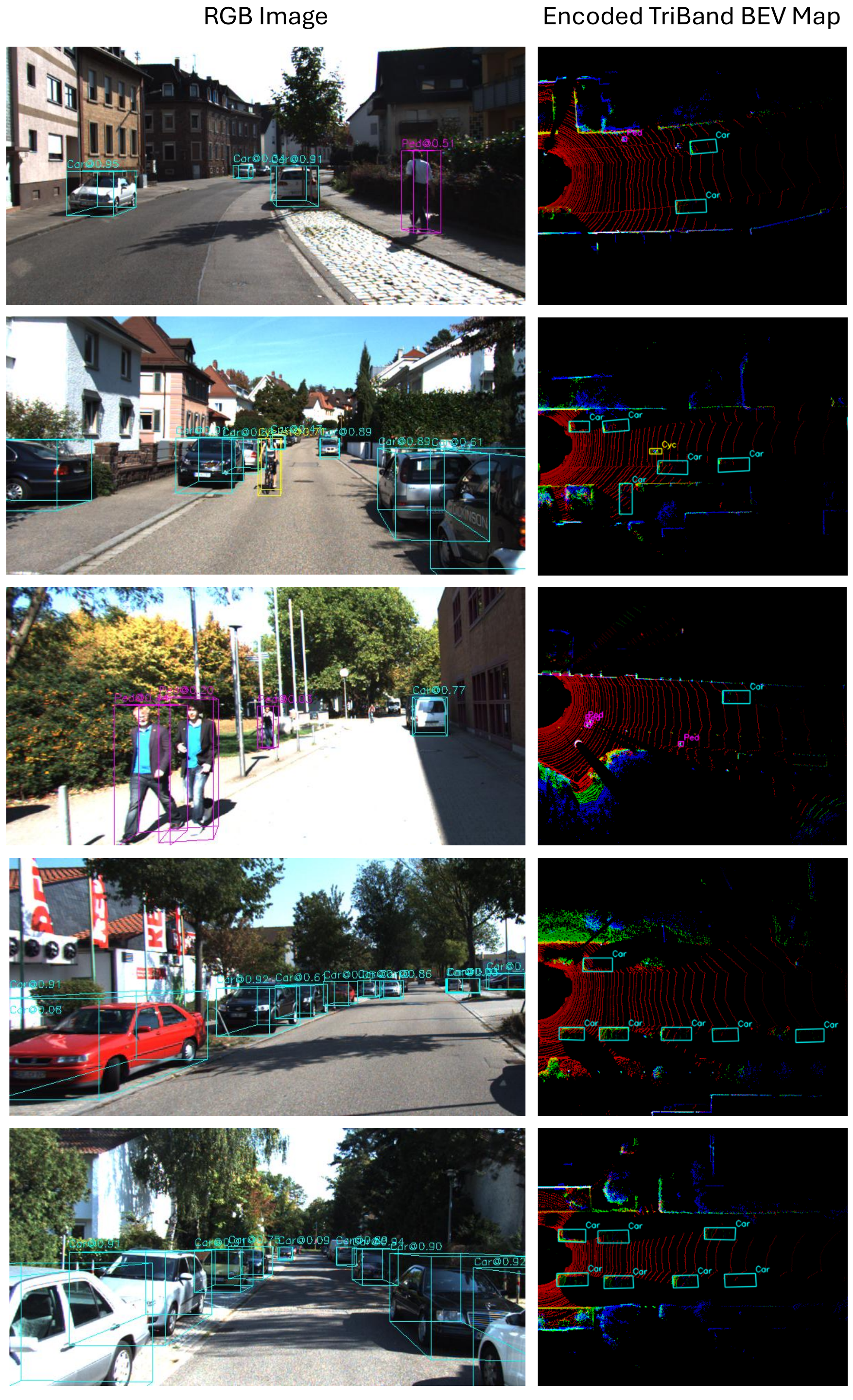}
  \caption{Qualitative detection results on validation scenes. The right side displays the TriBand encoded BEV map, and the left side shows the corresponding real camera image with transformed 3D predictions. Detections for pedestrian are shown in purple, car in cyan, and cyclist in yellow. The results demonstrate reliable handling of occlusions as well.}
  \label{fig:qual}
\end{figure}

Figure~\ref{fig:qual} visualizes representative validation scenes in camera and BEV map. Note that images are cropped to save space, so some distant predictions visible in the camera view may fall outside the BEV crop and thus not appear there. The detector localizes partially occluded objects with well aligned oriented boxes and avoids common BEV confounders such as slender roadside fixtures. The three band height encoding preserves vertical structure, so pedestrians often activate all three bands while short poles typically trigger a single narrow band. Layerwise silhouettes of legs, torso, and head also aid separability, yielding consistently low pedestrian false positives near curbside clutter.

\section{Discussion}
\label{sec:discussion}

Our achieved goal was to retain real time inference using a compact 2D BEV input while improving pedestrian detection performance. The pipeline runs at 49 FPS and the high resolution feature maps injected into the bidirectional fusion pathway further strengthen recall and localization for pedestrians. As discussed in Sec.~\ref{sec:exp:results}, horizontal localization is reliable, which suggests that the remaining limitation is height estimation rather than BEV detection quality. A practical next step is to pair the detector with a lightweight height predictor that refines the vertical extent only within the predicted object footprints in LiDAR space, keeping computation focused while improving 3D precision.

We tested global ground modeling using RANSAC and implemented GNDNet~\cite{paigwar2020gndnet}. Both added substantial latency because they process full 3D point clouds, whereas the local IQR-based compensation used here achieves stable results at much lower cost.

Increasing capacity to $C_{\text{base}}{=}64$ yielded only marginal AP gains (+1.1\% averaged for all classes) but raised compute by $2.8\times$ GFLOPS and increased memory demand. Such settings are not practical for real-time operation and offer diminishing returns relative to the proposed design.

\section{Conclusion}
\label{sec:conclusion}

This work presented a LiDAR-only real-time object detection framework that formulates 3D detection as a 2D learning problem through the proposed TriBand-BEV representation. The encoding captures vertical structure efficiently and allows the detector to operate directly on compact 2D BEV maps. The network employs a backbone with area attention modules and an extended bidirectional fusion pathway that integrates high-resolution feature maps. This design proved particularly effective for small or partially represented objects such as pedestrians, leading to consistent improvements in both BEV and 3D accuracy.

On the KITTI validation benchmark, the method achieved BEV AP of 58.7\%, 52.6\% and 47.2\% for pedestrians under the easy, moderate, and hard difficulty levels, exceeding the Complex-YOLO results by +7.76\% mAP (mean AP gain of three difficulty levels). Our network also detects cars and cyclists and qualitative analysis confirmed strong robustness under occlusion and clutter.

Overall, the model operates at 49 frames per second on light-weight BEV inputs, offering an effective balance between accuracy and efficiency. These results demonstrate the potential of height-aware BEV encoding combined with high-resolution bidirectional fusion for real-time LiDAR perception in mobile robotics and autonomous driving applications.

\section*{Acknowledgements}

 This work has been supported, in part, by the KIT Future Fields Wild Ideas 2026 program project "WildRobot". Furthermore, this research is part of the "CulturalRoad project", which has received funding from the European Union under grant agreement No. 101147397.

 Some of the trainings were performed using the resources provided by the Gauss Center for Supercomputing e.V. (GCS) through the John von Neumann Institute for Computing (NIC). Specifically, we utilized the GCS Supercomputer JUWELS located at the Jülich Supercomputing Center (JSC).




\bibliographystyle{ACM-Reference-Format} 
\bibliography{sample}


\end{document}